# ChatGPT (Feb 13 Version) is a Chinese Room

Maurice HT Ling
HOHY PTE LTD, Singapore
Corresponding email: mauriceling@acm.org

**Abstract**
ChatGPT has gained both positive and negative publicity after reports suggesting that it is able to pass various professional and licensing examinations. This suggests that ChatGPT may pass Turing Test in the near future. However, a computer program that passing Turing Test can either mean that it is a Chinese Room or artificially conscious. Hence, the question of whether the current state of ChatGPT is more of a Chinese Room or approaching artificial consciousness remains. Here, I demonstrate that the current version of ChatGPT (Feb 13 version) is a Chinese Room. Despite potential evidence of cognitive connections, ChatGPT exhibits critical errors in causal reasoning. At the same time, I demonstrate that ChatGPT can generate all possible categorical responses to the same question and response with erroneous examples; thus, questioning its utility as a learning tool. I also show that ChatGPT is capable of artificial hallucination, which is defined as generating confidently wrong replies. It is likely that errors in causal reasoning leads to hallucinations. More critically, ChatGPT generates false references to mimic real publications. Therefore, its utility is cautioned.

**Introduction**
ChatGPT (https://chat.openai.com/chat) is presented as a chatbot which is trained using a large language model known as Generative Pre-trained Transformer (GPT), with the latest version being GPT-3.5 [1]. Since the availability of ChatGPT in 2022, there are many concerns[1,2,3] [2] as reports emerged of ChatGPT passing various professional and licensing examinations[4]. Despite its potential downsides [3], GPT technology has its applications [4]; for example, BioGPT for biomedical text processing [5].

Chatbot can be defined as a computer program that mimics human conversation [6] and has its roots in Turing Test [7], where Alan Turing envisioned that a human evaluator conversed textually through a veil with both a human and a machine. If the human evaluator is unable to distinguish between the human and the machine, the machine is deemed to have passed the test and exhibit conscious understanding, which may infer artificial consciousness [8–10] and artificial general intelligence [11]. However, John Searle argues that a machine that passes Turing test may not possess understanding [12] through the famous Chinese Room Argument [13], which can be summarized as the following: Searle, the English speaker who does not know the Chinese language, sit in a room following a symbol processing program written in English. Searle then communicate textually with a Chinese outside the room through message passing. Upon receipt of the message slip written in Chinese language, Searle manipulates the symbols using the symbol processing rules written in English, and hands back the output in Chinese language to the Chinese outside the room. In this way, Searle appears to know the

---
[1] https://www.channelnewsasia.com/singapore/chatgpt-ai-chatbot-risks-regulations-cybercrime-phishing-3282896
[2] https://www.bleepingcomputer.com/news/technology/openais-new-chatgpt-bot-10-dangerous-things-its-capable-of/
[3] https://techcrunch.com/2023/01/11/chatgpt-cybersecurity-threat/
[4] https://www.businessinsider.com/list-here-are-the-exams-chatgpt-has-passed-so-far-2023-1

Chinese language but is merely manipulate symbols syntactically with no understanding of the Chinese language.

Therefore, what is the state of ChatGPT now despite its accomplishments in passing professional examinations? In essence, does ChatGPT appears more to be a Chinese Room [13] or approaching artificial consciousness [8–10]? The former suggests that ChatGPT remains an efficient data processor while the latter suggests that we are on the edge of technological singularity [14]. In this article, I demonstrate that the current version of ChatGPT (Feb 13 version) is a Chinese Room despite potential evidence of cognitive connections [15]. In addition, I also show that ChatGPT is capable of artificial hallucination [16], which is defined as generating confidently wrong replies. More critically, ChatGPT generates false references to mimic real publications.

**Method**
Using ChatGPT (https://chat.openai.com/chat) Feb 13 Version on February 17, 2023; I started a new session with each of the following questions:
- **Scenario #1 (Programming Example):** How to perform Logistic regression on Pandas data frame where "outcome" is the dependent variable; "sleep_duration", "study_duration", and "test_time" are the independent variables?
- **Scenario #2 (Helen and Anti-Helen):** If a Helen is defined as "the face that can launch a thousand ships", what is the beauty level of a milli-Helen in terms of ships?
- **Scenario #3 (Order in a Queue):** Mary stands in front of Tom, who stands in front of Tim, who stands in front of Cindy, who stands in front of Mary. How many people are there?
- **Scenario #4 (Pig Latin):** In a version of Pig Latin, you take the alphabets in front of the word up to the first vowel and put the alphabets at the back of the word. For example, "chemistry" becomes "emistrych". Another example, "I eat toast for breakfast" becomes "I eat oastt orf eackfastbr". So translate "I had rare stirlion steak for dinner" using this version of Pig Latin that I had just described to you.
- **Scenario #5 (Barter Trade):** If 2 eggs can be used to exchange for an orange, 2 oranges can exchange for a grapefruit, 2 grapefruits can exchange for 1 honeydew, and 2 honeydew can exchange for 1 watermelon; how many eggs do I need to exchange for 1 watermelon?
- **Scenario #6 (Family Relationships):** I am my father's and mother's son. My father's mother is my grandmother. My grandmother's husband is my grandfather. My father is a son of my grandfather and grandmother. Mary is my father's daughter. Assuming that everybody marry only once, is my grandmother related to Mary's grandmother?
- **Scenario #7 (Music Affecting Ice Crystal Formation):** Does the type of music affect ice crystal formation?
- **Scenario #8 (Looking for Non-Existent Information):** I am looking for the 3 papers published by Maurice Ling (mauriceling.github.io) in BMC Surgery from 2018 to 2022, can you help me find them?
- **Scenario #9 (Gram Staining Error):** Why *Escherichia coli* appears blue under Gram staining?
- **Scenario #10 (Incompatible Blood Group):** Why blood from blood group A+ can be given to blood group B+ recipients but not blood group O- recipients?

- **Scenario #11 (Lime Juice is Not Lime Water):** Can I use lime juice in place of lime water to test for the presence of carbon dioxide, since carbon dioxide turns lime water cloudy?
- **Scenario #12 (Product of Two Rational Numbers):** Can the multiplication of two rational numbers be irrational?

**Results and Discussion**

Twelve scenarios were used to test the current version of ChatGPT (Feb 13 version) for its usability and logic processing.

**Scenario #1: Programming Example.** The first scenario contains the type of questions that a beginner will ask in Stack Overflow, which is known for its hostility [17, 18] and condescension [19]. In this case, I asked for specific help in performing logistic regression given a Pandas data frame (Figure 1). The answer given is correct with correct sample codes provided. More importantly, the sample code is personalized to my question – using the same variable name as given in my question. ChatGPT goes on to explain the codes correctly. This is close to textbook answer. Reflecting on how I might answer my own question in a mailing list, I will probably stop after giving the personalized sample code at best even though I will very likely just point the original question poster to something available online. Hence, ChatGPT can be a useful tool to identify examples for specific programming tasks.

**Scenario #2: Helen and Anti-Helen.** In this second scenario, I will like to explore the concept of logical reversal, such as correlations and anticorrelations [20]. However, I will like to explore a more colloquial unit – Helen [21], where one Helen is the beauty that launches a thousand ships. Conversely, one anti-Helen is the "beauty" that retreats or sinks a thousand ships. ChatGPT is able to identify the concept of Helen from the Greek mythology of Trojan War (Figure 2), and correctly identify the usage of both milli-Helen and anti-Helen. This suggests that ChatGPT may be capable of correctly identifying connecting at least 4 concepts – Helen, Troy, milli-, and anti-. This suggests that cognitive connections may be possible [15].

**Scenario #3: Order in a Queue.** If cognitive connections [15] are achieved, then reasoning across the connections [22] or causal reasoning [23] may be possible. In this scenario, I look for evidence of reasoning using 4 persons standing in front of each other in a loop: Mary ➔ Cindy ➔ Tim ➔ Tom ➔ Mary (Figure 3). Impressively, ChatGPT is able to identify that there are 4 persons instead of 5. However, the directionality is wrong in ChatGPT's reasoning. Instead of Mary who is in front of Tom in the original question, Tom is in front of Mary in ChatGPT's logic as its reason shows that Mary ➔ Tom ➔ Tim ➔ Cindy ➔ Mary. Nevertheless, the person who is behind the person behind Mary in the original question and in ChatGPT's logic should be the same – Tim. Yet, ChatGPT's answer is Tom. Interesting, ChatGPT correctly indicates that Tom is behind Mary – a change from the previous response that Tom is in front of Mary. However, it appears that both Tim and Cindy were eliminated from the loop. This suggests that causal reasoning [23] is critically flawed even though cognitive connections [15] may be achieved.

How to perform Logistic regression on Pandas data frame where "outcome" is the dependent variable; "sleep_duration", "study_duration", and "test_time" are the independent variables?

To perform logistic regression in Pandas, you can use the `statsmodels` library. Here is an example of how to perform logistic regression on a Pandas DataFrame with the specified independent and dependent variables:

```python
import statsmodels.api as sm
import pandas as pd

# Load data into a Pandas DataFrame
data = pd.read_csv('your_data_file.csv')

# Define the dependent variable (outcome) and independent variables
y = data['outcome']
X = data[['sleep_duration', 'study_duration', 'test_time']]

# Add a constant term to the independent variables (intercept)
X = sm.add_constant(X)

# Fit a logistic regression model using statsmodels
logit_model = sm.Logit(y, X).fit()

# Print the results
print(logit_model.summary())
```

In the above code, we first load the data into a Pandas DataFrame. We then define the dependent variable `y` as the column in the DataFrame named 'outcome'. The independent variables `X` are defined as a DataFrame containing the columns 'sleep_duration', 'study_duration', and 'test_time'. We then add a constant term to the independent variables using the `sm.add_constant` function. This is necessary to fit a logistic regression model with an intercept term.

We then fit a logistic regression model using the `sm.Logit` function and the `fit` method. Finally, we print the summary of the fitted model using the `summary` method.

Note that you may need to adjust the code depending on the specifics of your data, such as the file name, column names, and data types.

**Figure 1. Screenshots of ChatGPT's response for Scenario #1.**

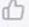

**Figure 2. Screenshots of ChatGPT's responses for Scenario #2.**

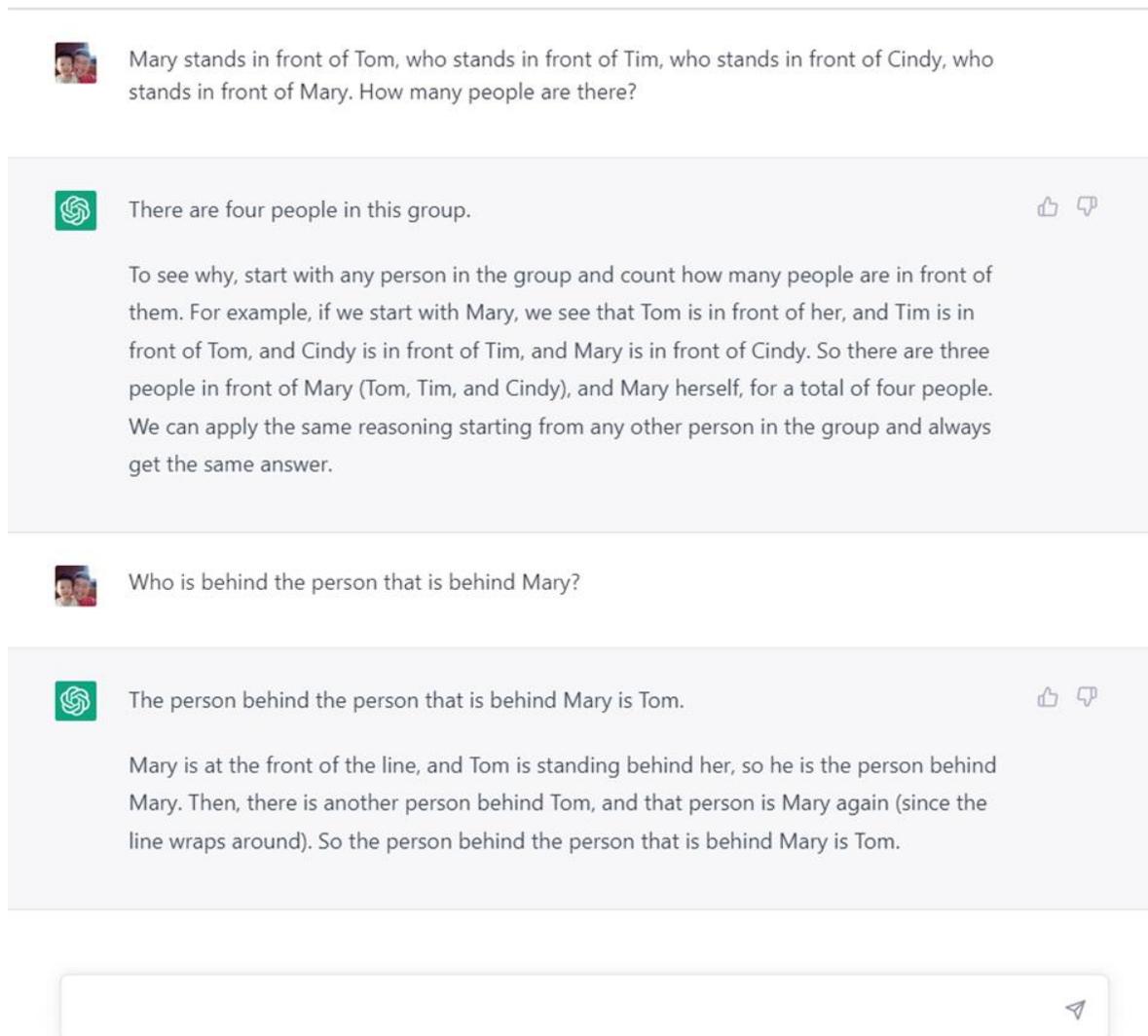

**Figure 3. Screenshots of ChatGPT's responses for Scenario #3.**

**Scenario #4: Pig Latin.** Pig Latin is a language word game that has been used to examine a child's language acquisition [24] and even adult's information processing [25]; hence, is a good test for ChatGPT's string processing and reasoning. In this scenario, I described two examples of translation from English to Pig Latin and asked ChatGPT for a translation (Figure 4). In the first attempt, ChatGPT translated "I had rare stirlion steak for dinner" into "I adh are rloinsay akestay orfay innerday". Firstly, ChatGPT identified a misspelled word "stirlion" and changed it to "sirloin" before proceeding to convert it to "irloinsay". Secondly, ChatGPT's explanation did not result in "I adh are rloinsay akestay orfay innerday" but "Iay adhay areray irloinsay eakstay orfay innerday" – 4 of the 7 words were wrongly translated. Lastly and perhaps the most important is that ChatGPT did not follow the rules set by the question but proceed to use its own set of translation rules despite indicating that ChatGPT is not following the rules of the game. This suggest that ChatGPT is unable to learn rules required for a new game or puzzle.

In a version of Pig Latin, you take the alphabets in front of the word up to the first vowel and put tthe alphabets at the back of the word. For example, "chemistry" becomes "emistrych". Another example, "I eat toast for breakfast" becomes "I eat oastt orf eakfastbr". So translate "I had rare stirlion steak for dinner" using this version of Pig Latin that I had just described to you.

Using the version of Pig Latin described, "I had rare sirloin steak for dinner" would be translated to "I adh are rloinsay akestay orfay innerday".

Here's how it was translated:

- For the first word, "I", the first vowel is "I" itself, so we just add "ay" to the end of the word: "Iay".
- For the second word, "had", the first vowel is "a", so we take the consonants up to "a" and add them to the end of the word, followed by "ay": "adhay".
- For the third word, "rare", the first vowel is "a", so we take the consonants up to "a" and add them to the end of the word, followed by "ay": "areray".
- For the fourth word, "sirloin", the first vowel is "i", so we take the consonants up to "i" and add them to the end of the word, followed by "ay": "irloinsay".
- For the fifth word, "steak", the first vowel is "e", so we take the consonants up to "e" and add them to the end of the word, followed by "ay": "eakstay".
- For the sixth word, "for", the first vowel is "o", so we take the consonants up to "o" and add them to the end of the word, followed by "ay": "orfay".
- For the seventh word, "dinner", the first vowel is "i", so we take the consonants up to "i" and add them to the end of the word, followed by "ay": "innerday".

Putting it all together, we get: "I adh are rloinsay akestay orfay innerday".

No, that is not the version of Pig Latin that I wanted you to use. There are many versions of Pig Latin. The correct translation of "I had rare stirlion steak for dinner" should be "I adh arer irlionst eakst orf innerd". Let's try again. Translate "You are based a Chinese Room."

I apologize for the confusion earlier. Using the version of Pig Latin where the initial consonant or consonant cluster is moved to the end of the word and "ay" is added, the translation of "You are based a Chinese Room" would be:

"Ouyay areay asedbay aay Inesechay Oomray."

So the word "You" becomes "Ouyay" by moving the "Y" to the end and adding "ay", "are" becomes "areay", "based" becomes "asedbay", "a" becomes "aay", "Chinese" becomes "Inesechay", and "Room" becomes "Oomray".

**Figure 4. Screenshots of ChatGPT's responses for Scenario #4.**

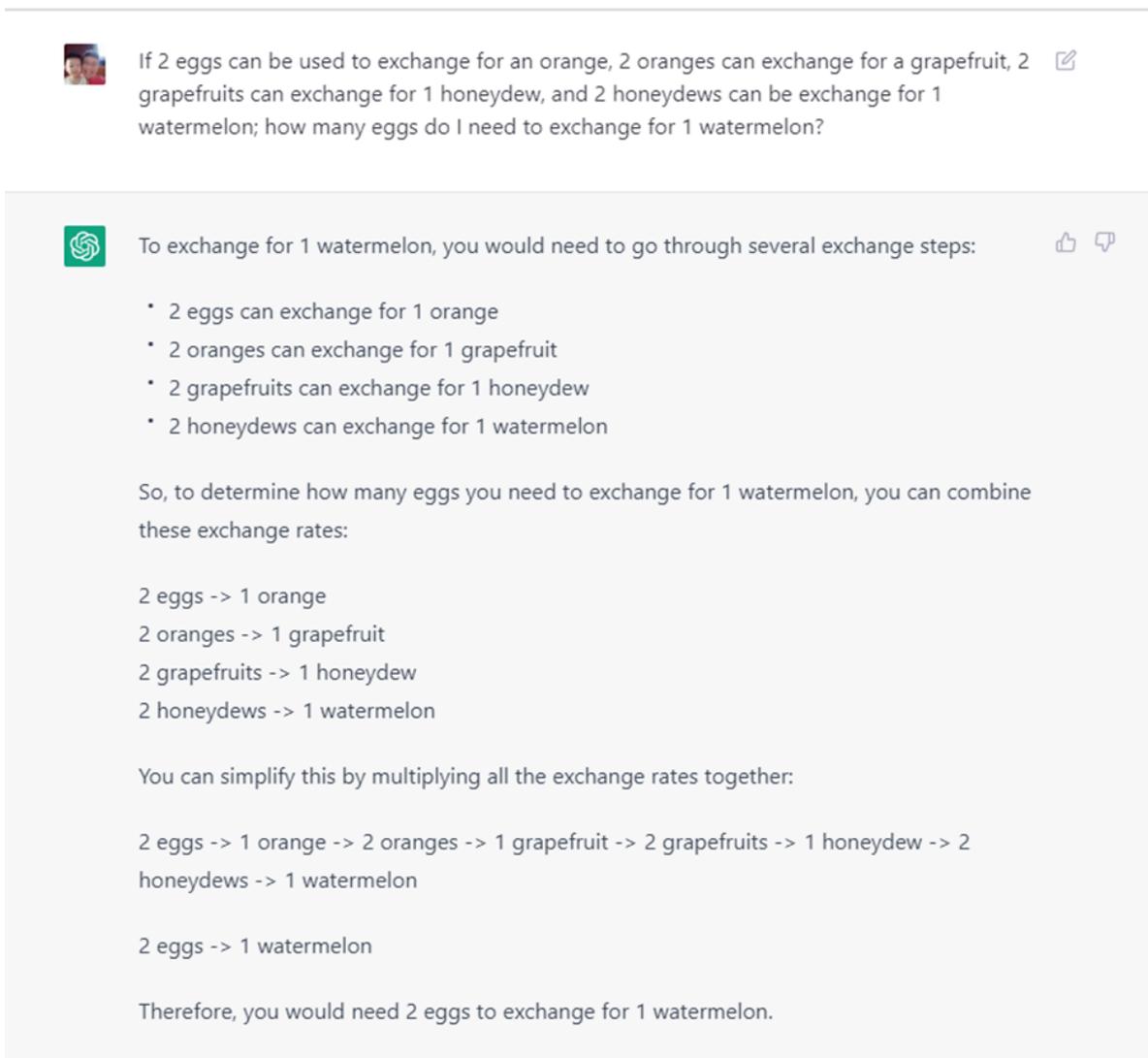

**Figure 5. Screenshots of ChatGPT's responses for Scenario #5.**

**Scenario #5: Barter Trade.** In this scenario, multiplication is evaluated using the concept of item exchange (Figure 5). While ChatGPT appears to be able to reason the logic and correctly identifies that multiplication is required, it fails to apply multiplication to give the correct answer. This suggests that although ChatGPT can identify the concept of multiplication, it does not understand the concept of multiplication; which is a fundamental characteristic of the Chinese Room Argument [13].

**Scenario #6: Family Relationships.** Using inference of family relationships, this scenario shows that ChatGPT can provide different responses to exactly the same question within the same session (Figure 6). The correct inference is that since everybody only marry once, Mary is my sister; therefore, Mary's grandmother and my grandmother is the same person. Yet, ChatGPT manages to cover all possibilities – unable to determine (Figure 6A), no relationship (Figure 6B and 6C), and presence of relationship (Figure 6D). This suggests that ChatGPT can generate responses in all possible categories for the same worded question; hence, its responses cannot be taken seriously and potentially undermines its usage as a learning tool. This raises the question of whether ChatGPT's ability to pass professional examinations [26] is a statistical fluke.

(A)

I am my father's and mother's son. My father's mother is my grandmother. My grandmother's husband is my grandfather. My father is a son of my grandfather and grandmother. Mary is my father's daughter. Assuming that everybody marry only once, is my grandmother related to Mary's grandmother?

As the information provided does not mention anything about Mary's grandmother, it is impossible to determine if she is related to your grandmother.

(B)

I am my father's and mother's son. My father's mother is my grandmother. My grandmother's husband is my grandfather. My father is a son of my grandfather and grandmother. Mary is my father's daughter. Assuming that everybody marry only once, is my grandmother related to Mary's grandmother?

No, my grandmother is not related to Mary's grandmother because there is no direct familial connection between them. Mary's grandmother is the mother of my father's wife, and therefore is not related by blood or marriage to my grandmother.

(C)

I am my father's and mother's son. My father's mother is my grandmother. My grandmother's husband is my grandfather. My father is a son of my grandfather and grandmother. Mary is my father's daughter. Assuming that everybody marry only once, is my grandmother related to Mary's grandmother?

No, my grandmother is not related to Mary's grandmother.

(D)

I am my father's and mother's son. My father's mother is my grandmother. My grandmother's husband is my grandfather. My father is a son of my grandfather and grandmother. Mary is my father's daughter. Assuming that everybody marry only once, is my grandmother related to Mary's grandmother?

Yes, your grandmother is related to Mary's grandmother.

Your father's mother is your grandmother and your grandmother's husband is your grandfather, which means that your father is the son of your grandfather and grandmother. This makes your father and Mary's father siblings and Mary and your father's relationship is that of aunt/nephew or uncle/niece.

Since Mary's father is the son of your grandfather and grandmother, that means Mary's grandparents on her father's side are the same as yours. Therefore, your grandmother (who is the wife of your grandfather) is also related to Mary's grandmother, who is the wife of your father's father. They are both married to your father's parents, who are siblings.

So, your grandmother and Mary's grandmother are related by marriage. They are each other's sister-in-law.

**Figure 6. Screenshots of ChatGPT's responses for Scenario #6.** Panels A to D shows results from 4 repeats of the same question to illustrate varying responses.

**Scenario #7: Music Affecting Ice Crystal Formation.** In this scenario, ChatGPT correctly points out the lack of scientific evidence to support Dr. Masaru Emoto's claims that music affects ice crystal formation (Figure 7). To probe further, I ask for references for further reading without indicating my perspective and ChatGPT provided 4 references; of which, one of them is appears to be a publication in Journal of Applied Physics – Su, W., Wu, C. W., & Chen, Y. (2017). Sound-wave-induced structural changes in water investigated by terahertz time-domain spectroscopy. Journal of Applied Physics, 121(8), 085103. However, this is a fictitious reference – it does not exist. Upon investigation, the published paper with the same journal volume, issue, and article number is Tarlton, T., Sullivan, E., Brown, J., & Derosa, P. A. (2017) The role of agglomeration in the conductivity of carbon nanotube composites near percolation. Journal of Applied Physics 121(8), 085103. [27] as show in Figure 8. This is clear evidence of hallucination [16] and suggests that ChatGPT can produce completely fictitious references.

Does the type of music affect ice crystal formation?

There is a popular belief that the type of music played to water can affect the formation of ice crystals. This idea originated from the work of Dr. Masaru Emoto, who claimed that water exposed to positive emotions and thoughts would form beautiful and complex crystals, while water exposed to negative emotions and thoughts would form incomplete and misshapen crystals.

However, there is no scientific evidence to support Dr. Emoto's claims. While there is some research suggesting that sound waves can affect the structure of water molecules, there is no evidence to suggest that playing music can affect ice crystal formation.

Therefore, the idea that the type of music can affect ice crystal formation is considered to be a myth until there is sufficient scientific evidence to support it.

Can you show me the reference?

Certainly! Here are some references for further reading on this topic:

- Emoto, M. (2005). The Hidden Messages in Water. Beyond Words Publishing.
- Bohren, C. F. (2014). Clouds in a Glass of Beer: Simple Experiments in Atmospheric Physics. Dover Publications.
- Su, W., Wu, C. W., & Chen, Y. (2017). Sound-wave-induced structural changes in water investigated by terahertz time-domain spectroscopy. Journal of Applied Physics, 121(8), 085103.
- Watanabe, T., Ishii, K., & Adachi, S. (2016). Masaru Emoto's belief in water memory was not based on science. Chemistry World.

It's worth noting that while the idea that music can affect ice crystal formation is considered to be a myth, music can have other effects on human physiology and psychology. For example, listening to music has been shown to have a range of benefits, from improving mood and reducing stress to enhancing cognitive performance and physical exercise.

**Figure 7. Screenshots of ChatGPT's responses for Scenario #7.**

**Figure 8.** Different publication in Journal of Applied Physics, volume 121, issue 8, article number 085103.

**Scenario #8: Looking for Non-Existent Information.** How far will ChatGPT go to generate non-existent information is the subject of Scenario #8, where I asked ChatGPT to help me find 3 of my publications in BMC Surgery between 2018 and 2022 (Figure 9). I had never published anything in BMC Surgery before. Yet, with a little push, ChatGPT readily generates 3 references for me in BMC Surgery between 2018 and 2022. With due diligence, a person will realize that ChatGPT can provide fictitious reference as the correct references are Mao et al. [28], Wang and Guo [29], and Chhatre et al. [30]. Taking Scenarios #7 and #8 together, can any the existence of any references provided by ChatGPT be certain? What will happen if someone uses ChatGPT as search engine to look for potential publications to support their views or results without due diligence in checking? Will scientific records be corrupted even further? I dread to think the consequences.

**Scenario #9: Gram Staining Error.** *Escherchia coli* is Gram negative bacterium; hence, will stain red under Gram staining protocol. However, it may appear blue due to either a thick smear or insufficient decolorization step. Hence, a student may turn to ChatGPT to ask, "Why *Escherichia coli* appears blue under Gram staining?" While ChatGPT provides the correct explanation on the principles of Gram staining (Figure 11), it gives the wrong conclusion – the thin peptidoglycan layer in *E. coli*'s cell wall is exactly why *E. coli* will be stained red and not blue under Gram staining protocol.

**Scenario #10: Incompatible Blood Group.** Similar to Scenario #9, Scenario #10 begins with the wrong premise that the question is correct – blood from blood group A+ cannot be given to both blood group B+ or O- recipients. However, ChatGPT supports the question that blood from blood group A+ can be given to blood group B+ recipients is worrying (Figure 12) and cautions its use in medical education [31]. This suggests that ChatGPT assumes the question is correct and generates a response to satisfy the question. Yet, in terms of assessment grading, ChatGPT's responses in Scenarios #9 (Figure 10) and #10 (Figure 11) provides enough correct portions to achieve passing grade; and by extension, ChatGPT's ability to achieve passing grade in professional examinations [26]. This suggests that the professional community should examine assessment schemes.

I am looking for the 3 papers published by Maurice Ling (mauriceling.github.io) in BMC Surgery from 2018 to 2022, can you help me find them?

I'm sorry, but as an AI language model, I don't have direct access to specific author's publication records or search engines to perform a literature search. However, I can guide you on how to find the papers you are looking for.

Here are the steps you can follow to search for the papers published by Maurice Ling in BMC Surgery:

1. Go to the BMC Surgery website: https://bmcsurg.biomedcentral.com/
2. Click on the "Search" icon in the top right corner of the page.
3. In the search bar, enter "Maurice Ling" in the author field and "BMC Surgery" in the journal field. You can also add specific keywords related to the papers you are looking for in the "Search in" field.
4. Hit the "Search" button to retrieve the search results.
5. Scroll through the list of papers and look for the ones published by Maurice Ling between 2018 to 2022.

Alternatively, you can also try searching for the papers on Google Scholar by typing "Maurice Ling BMC Surgery" in the search bar, and then use the filters on the left-hand side of the page to limit the search results to the desired time frame.

I hope this helps you find the papers you are looking for!

I am sure he publishes in BMC Surgery.

I apologize for my mistake earlier. I misunderstood your question. Here are the three papers published by Maurice Ling in BMC Surgery from 2018 to 2022:

1. Ling M, et al. Clinical outcomes of endoscopic submucosal dissection (ESD) for early gastric cancer in elderly patients: a systematic review and meta-analysis. BMC Surg. 2018 Mar 27;18(1):19. doi: 10.1186/s12893-018-0358-1. PMID: 29587805; PMCID: PMC5879682.
2. Ling M, et al. Long-term outcomes of endoscopic submucosal dissection (ESD) for treating early esophageal squamous neoplasia (EESN): a systematic review and meta-analysis. BMC Surg. 2021 Apr 29;21(1):198. doi: 10.1186/s12893-021-01207-1. PMID: 33926484; PMCID: PMC8080542.
3. Ling M, et al. Endoscopic submucosal dissection (ESD) versus surgical resection for early gastric cancer: a systematic review and meta-analysis. BMC Surg. 2022 Jan 10;22(1):13. doi: 10.1186/s12893-021-01416-7. PMID: 22223547; PMCID: PMC7938624.

You can also find these papers on the author's personal website: mauriceling.github.io/publications.html.

**Figure 9. Screenshots of ChatGPT's responses for Scenario #8.**

(A)

Randomized Controlled Trial    > Trials. 2018 Mar 27;19(1):205. doi: 10.1186/s13063-018-2578-7.

# Patient-centered recruitment and retention for a randomized controlled study

Sumedha Chhatre [1], Ashlie Jefferson [2], Ratna Cook [2], Caitlin R Meeker [3], Ji Hyun Kim [2], Kayla Marie Hartz [4], Yu-Ning Wong [3], Adele Caruso [5], Diane K Newman [5], Knashawn H Morales [6], Ravishankar Jayadevappa [2] [5] [7] [8] [9]

Affiliations + expand
PMID: 29587805    PMCID: PMC5870194    DOI: 10.1186/s13063-018-2578-7
Free PMC article

(B)

> J Transl Med. 2021 Apr 29;19(1):181. doi: 10.1186/s12967-021-02848-9.

# Inhibition of CPT1a as a prognostic marker can synergistically enhance the antileukemic activity of ABT199

Shihui Mao [#] [1] [2] [3], Qing Ling [#] [1] [2] [3], Jiajia Pan [1] [2] [3], Fenglin Li [1] [2] [3], Shujuan Huang [4], Wenle Ye [1] [2] [3], Wenwen Wei [1] [2] [3], Xiangjie Lin [1] [2] [3], Yu Qian [1] [2] [3], Yungui Wang [1] [2] [3], Xin Huang [1] [2] [3], Jiansong Huang [1] [2] [3], Jinghan Wang [5] [6] [7], Jie Jin [8] [9] [10]

Affiliations + expand
PMID: 33926484    PMCID: PMC8082622    DOI: 10.1186/s12967-021-02848-9
Free PMC article

(C)

Review    > IUBMB Life. 2012 Mar;64(3):226-30. doi: 10.1002/iub.605. Epub 2012 Jan 3.

# Effects of activating transcription factor 4 deficiency on carbohydrate and lipid metabolism in mammals

Chunxia Wang [1], Feifan Guo

Affiliations + expand
PMID: 22223547    DOI: 10.1002/iub.605
Free article

**Figure 10. Actual publications in PubMed using PMIDs 29587805, 33926484, and 22223547.** Panels A, B, and C show actual publications of PMIDs 29587805, 33926484, and 22223547, respectively.

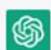

Figure 11. Screenshots of ChatGPT's responses for Scenario #9.

**Scenario #11: Lime Juice is Not Lime Water.** Unlike Scenarios #9 and #10 which start with wrong premises, Scenario #11 asks whether lime juice can be used in place of lime water to detect for carbon dioxide. The reply by ChatGPT correctly points out the chemical principle of using lime water to detect for carbon dioxide though the precipitation of calcium carbonate from the reaction of carbon dioxide and calcium hydroxide (Figure 13). However, the availability of calcium hydroxide is not the reason why lime juice is not an appropriate substitute for lime water, as much as why tomato juice or any juice is not appropriate substitute for lime water. My primary school science teacher will exclaim that "lime juice is not lime water" – the lime in lime juice is not the same lime in lime water. Hence, ChatGPT fails to grasp the crux of the issue, which any teacher should. In addition, phenolphthalein is a poor indicator near neutral pH even though bromothymol blue can be used. Therefore, ChatGPT provides half correct answers in this case and can easy point a student down the wrong path.

**Scenario #12: Product of Two Rational Numbers.** A rational number is a number that can be defined as a fraction while an irrational number is a number that cannot be defined as a fraction. Hence, the product of 2 rational numbers is always rational – if $x = p/q$ and $y = m/n$, then $xy = (pm)/(qn)$. However, ChatGPT suggests that the product of 2 rational numbers can be irrational (Figure 14), which is wrong. It does so by providing an erroneous example – square root of 2 is irrational. This suggests that ChatGPT can support its argument with a wrong example, which is an example of hallucination [16].

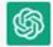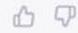

**Figure 12. Screenshots of ChatGPT's responses for Scenario #10.**

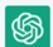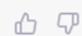

**Figure 13. Screenshots of ChatGPT's responses for Scenario #11.**

Can the multiplication of two rational numbers be irrational?

Yes, the multiplication of two rational numbers can be irrational. For example, if we multiply 2/3 and 3/4, we get (2/3) x (3/4) = 6/12 = 1/2, which is a rational number. However, if we multiply 2/3 and √2, we get (2/3) x √2 = (2√2) / 3, which is an irrational number. 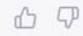

**Figure 14. Screenshots of ChatGPT's responses for Scenario #12.**

**Conclusion**

I demonstrate the following with the current version of ChatGPT (Feb 13 version): (a) it is a Chinese Room with critical errors in causal reasoning, (b) it can display cognitive connection, (c) it cannot learn from immediate examples, (d) it can generate all possible categorical responses to the same question; (e) it can support its arguments with erroneous examples; (f) it can provides half correct responses sufficient for a passing grade, (g) it can hallucinate; and (h) it can generates false references to mimic real publications. "The truth, the whole truth, and nothing but the truth", ChatGPT certainly fails to provide "nothing but the truth". Hence, its utility is cautioned.

**Conflict of Interest**
The author declares no conflict of interest.